\documentclass[a4paper]{article}

\usepackage{INTERSPEECH2022}
\usepackage{amsthm}
\usepackage{comment, url}

\title{Bayesian Recurrent Units and the Forward-Backward Algorithm}

\name{Alexandre Bittar$^{\,1,2}$, Philip N. Garner$^{\,1}$}
\address{
    $^1$Idiap Research Institute, Martigny, Switzerland \\
    $^2$\'Ecole Polytechnique Fédérale de Lausanne, Switzerland
    }
\email{abittar@idiap.ch, pgarner@idiap.ch}

\begin{document}

\maketitle
%%%%%%%%%% ABSTRACT %%%%%%%%%% 
\begin{abstract}
    Using Bayes's theorem, we derive a unit-wise recurrence as well as a backward recursion similar to the forward-backward algorithm. The resulting Bayesian recurrent units can be integrated as recurrent neural networks within deep learning frameworks, while retaining a probabilistic interpretation from the direct correspondence with hidden Markov models. Whilst the contribution is mainly theoretical, experiments on speech recognition indicate that adding the derived units at the end of state-of-the-art recurrent architectures can improve the performance at a very low cost in terms of trainable parameters. 
\end{abstract}

%%%%%%%%%% 1.) INTRO %%%%%%%%%%
\section{Introduction}

Recurrent models have been widely employed in signal processing and statistical pattern recognition, notably in the form of Kalman's state space filter \cite{Kalman1960, Scharf1991} and hidden Markov models (HMMs) \cite{Baum1966, Baum1967, Baum1970, Bahl1983}. Both approaches use a forward-backward training procedure to make a statistical estimation of the model parameters.

With the current success of machine learning techniques, speech recognition architectures can be trained in an end-to-end fashion, by exploiting auto-differentiation inside deep learning frameworks like PyTorch \cite{Paszke2017}. Here, recurrence is also an important concept and recurrent neural networks (RNNs) are commonly trained via the error Back-Propagation Through Time (BPTT) algorithm \cite{Rumelhart1986, Williams1989}, which is a generalization of gradient descent to process sequential data. During the forward pass, a batch of input examples is passed through the network, and a loss function is applied to the final outputs. During the subsequent backward computation of derivatives, the network trainable parameters are updated to minimize the loss.

Similarities between HMMs and RNNs have long been observed. Bourlard and Wellenkens \cite{Bourlard1990} have shown that the outputs of RNNs approximate the maximum \textit{a posteriori} (MAP) output probabilities of HMMs trained via the Viterbi algorithm \cite{Forney1973}. Bridle \cite{Bridle1990} also showed that the \textit{alpha} part of the forward-backward algorithm can be simulated by a recurrent network, and that the \textit{beta} part bears similarities with the backward computation of derivatives in the training of neural networks. Bidirectional RNNs were subsequently defined by Schuster and Paliwal \cite{Schuster1997} to explicitly allow networks to take into account future observations. This approach was later applied to gated RNNs \cite{Graves2005} and is now the standard approach for LSTMs \cite{Hochreiter1997}, GRUs \cite{Cho2014} and Li-GRUs \cite{Ravanelli2018}. In a bidirectional recurrent layer, the size of all feedforward weight matrices (i.e., matrices that are applied to the layer inputs) is doubled compared to the unidirectional case, which represents a significant increase in the amount of trainable parameters. Recently, Garner and Tong \cite{Garner2021} have used a Bayesian interpretation of gated RNNs to derive a backward recursion through the input sequences. Their probabilistic approach, analogous to a Kalman smoother, allows the consideration of future observations without requiring any additional trainable parameters. 

This Bayesian approach treats the input as a sequence of observations, and interprets the unit outputs as the probabilities of hidden features being present at each timestep. Recurrence emerges naturally from Bayes's theorem which updates a prior probability into a posterior given new observational data. In our previous work on the light Bayesian recurrent unit \cite{Bittar2021}, hidden features were assumed to be interdependent, which led to a layer-wise recurrence for the computation of prior probabilities. In this paper, in a mainly theoretical contribution, we come back to the simpler case of a RNN with unit-wise recurrence and no gate. Here, by assuming a latent space of independent features, a Bayesian analysis demonstrates that the trainable parameters of the network directly correspond to standard parameters of a first-order 2-state hidden Markov model (HMM). 

In a second step, similarly to the Kalman smoother \cite{Kalman1960} and to the forward-backward algorithm \cite{Baum1972}, we derive two different backward recursions that allow the consideration of future observations without relying on any additional parameters. We also prove by induction that the two are equivalent. In contrast with the work of Garner and Tong \cite{Garner2021}, the unit-wise recurrence is here derived using transition probabilities instead of a context relevance gate. 

The derived unit-wise Bayesian recurrent units (UBRUs) can be trained like standard RNNs inside a modern deep neural network (DNN). Even though UBRUs have much less representational power than state-of-the-art RNNs, they are appropriate when the features are decorrelated. We confirm this by showing that, when placed on the phoneme (rather than acoustic) region of a DNN for automatic speech recognition (ASR), they are able to replace larger standard bidirectional gated RNNs without any loss of performance. More generally, our approach aims at developing the growing toolkit of Bayesian techniques applicable to deep learning.

%%%%%%%%%% 2.) HMM APPROACH %%%%%%%%%%
\section{A hidden Markov model approach}

Consider an input sequence $\boldsymbol X_T=[\boldsymbol x_1,\hdots,\boldsymbol x_T]\in\mathbb{R}^{F\times T}$ of length $T$, where each observation $\boldsymbol x_t$ is a vector with $F$ input dimensions. We assume that there are $H$ hidden features $\{\phi_i\,| i=1,\hdots,H\}$ that we wish to detect along the sequence. At each timestep $t$, a feature has two possible states: present or absent, that we write as $\phi_{t,i}$ and $\neg\phi_{t,i}$ respectively. Each hidden feature can be represented as a first-order 2-state Markov process, such that its probability of occurring at timestep $t$ only depends on its state at the previous timestep $t-1$. For a single hidden feature $\phi$, an initial state probability $a\in[0,1]^{2\times 1}$ ,
\begin{equation}
a=\Big[P(\phi_{0}), P(\neg\phi_{0})\Big] \, ,
\end{equation}
and a transition matrix $A\in[0,1]^{2\times2}$,
\begin{equation}
A=
\begin{bmatrix}
P(\phi_{t}|\phi_{t-1}) &
P(\neg\phi_{t}|\phi_{t-1}) \\[5pt]
P(\phi_{t}|\neg\phi_{t-1}) & P(\neg\phi_{t}|\neg\phi_{t-1})
\end{bmatrix} \, ,
\end{equation}
can be defined to describe the evolution of the state through discrete time. Then, for any binary sequence of hidden states, the probability of the sequence being generated by the Markov chain is fully defined in terms of $a$ and $A$ as a product of initial and transition probabilities. Since the hidden sequence is not directly observable, let us additionally define a set of distributions, 
\begin{equation}
B(\boldsymbol x_t)=\Big[b_1(\boldsymbol x_t),b_2(\boldsymbol x_t) \Big]=\Big[p(\boldsymbol x_t|\phi_{t}),\,p(\boldsymbol x_t|\neg\phi_{t})\Big] \, ,
\end{equation}
representing the likelihood of seeing observation $\boldsymbol x_t$ at timestep $t$ given the two possible feature states. As explained by Juang and Rabiner \cite{Juang1991}, the stochastic process represented by $\boldsymbol X_T$ can then be fully characterized by the HMM parameters $a$, $A$ and $B(\boldsymbol x_t)$, without requiring the knowledge of the sequence of hidden states.

%%%% Machine Learning %%%%
\subsection{Neural network formulation}

Let us start by using a more machine learning oriented formulation of the HMM parameters $a$ and $A$. We define trainable scalars $\rho_{0,i}$, $\tau_{11,i}$ and $\tau_{01,i}$ $\in[0,1]$ that describe the initial and transition probabilities of the $i$-th hidden feature,
\begin{equation}
a_i=\Big[\rho_{0,i}, 1-\rho_{0,i}\Big] \quad\text{and}\quad A_i=
\begin{bmatrix}
\tau_{11,i} & 1-\tau_{11,i} \\
\tau_{01,i} & 1-\tau_{01,i}
\end{bmatrix} \, ,
\end{equation}
where we used the notation $\tau_{kl}=P(\phi_t=l|\phi_{t-1}=k)$, $k,l\in\{0,1\}$. These can then be vectorized for the whole layer as $\boldsymbol \rho_0$, $\boldsymbol \tau_{11}$ and $\boldsymbol \tau_{01}$ $\in[0,1]^H$. In order to express the remaining HMM parameters related to the set of distributions $\boldsymbol B(\boldsymbol x_t)$, we can assume that the likelihood of observing $\boldsymbol x_t$ given the current state of the hidden features $\boldsymbol\phi_t$, can be described using a distribution from the exponential family. As we will see in Section \ref{ssec:forward}, only the ratio of these distributions will be necessary to compute. As demonstrated by Garner and Tong \cite{Garner2021} drawing from Bridle \cite{Bridle1990b}, this ratio of likelihood $\boldsymbol r_t$ can then be expressed as
\begin{equation}
\label{eq_rt}
    \boldsymbol r_t:=\frac{p(\boldsymbol x_t|\,\neg\boldsymbol\phi_t\,)}{p(\boldsymbol x_t|\,\boldsymbol\phi_t\,)}=\exp\Big[-\boldsymbol W^T\,\boldsymbol x_t - \boldsymbol b \Big] \, .
\end{equation}
Similarly to \cite{Bittar2021}, for the case of multivariate normal distributions that share the same covariance matrix $\boldsymbol\Sigma$, i.e., $p(\boldsymbol x_t|\boldsymbol\phi_t)\sim\mathcal{N}(\boldsymbol\mu,\boldsymbol\Sigma)$ and $p(\boldsymbol x_t|\,\neg\boldsymbol\phi_t\,)\sim\mathcal{N}(\boldsymbol\nu,\boldsymbol\Sigma)$, the parameters $\boldsymbol W\in\mathbb{R}^{F\times H}$ and $\boldsymbol b\in\mathbb{R}^{H}$ can be expressed as,
\begin{subequations}
\begin{align}
    \boldsymbol W&=\Big(\boldsymbol\nu^T-\boldsymbol\mu^T\Big)\,\boldsymbol\Sigma^{-1} \\
    \boldsymbol b\, &= -\frac{1}{2}\Big(\boldsymbol\nu^T\,\boldsymbol\Sigma^{-1}\,\boldsymbol\nu+\boldsymbol\mu^T\,\boldsymbol\Sigma^{-1}\,\boldsymbol\mu\Big) \, .
\end{align}
\end{subequations}
Overall, we have shown that the Markov processes corresponding to a layer of $H$ independent hidden features can be fully described by a set of trainable tensors (or parameters) $\boldsymbol \rho_0$, $\boldsymbol \tau_{11}$, $\boldsymbol \tau_{01}$ $\in[0,1]^H$, $\boldsymbol W\in\mathbb{R}^{F\times H}$ and $\boldsymbol b\in\mathbb{R}^{H}$. In the next section, we will derive a forward-backward formulation that is similar to that of recurrent neural networks (RNNs). This will allow them to be trained inside a machine learning framework, while retaining a probabilistic interpretation as they correspond to standard HMM parameters.

%%%%%%%%%% 3.) FORWARD-BACKWARD %%%%%%%%%% 
\section{Forward-backward procedure}

In order to make inference about the state of the hidden features throughout the sequence, we use a Bayesian approach and design a layer of recurrent units that will evaluate the stacked conditional probabilities $\boldsymbol \gamma_t:=P(\boldsymbol\phi_t|\boldsymbol X_T)\in[0,1]^H$ of the different features being present at each timestep $t=1,\hdots,T$, given the information of the complete input sequence $\boldsymbol X_T$. In the first \textit{alpha} or forward part of the procedure, the probabilities $\boldsymbol \alpha_t:=P(\boldsymbol\phi_t|\boldsymbol X_t)\in[0,1]^H$ are computed. In the subsequent \textit{beta} or backward part, these probabilities are smoothed by taking into account future observations and produce the desired outputs $\boldsymbol \gamma\in[0,1]^{T\times H}$, that are fed into the next layer.

%%%% Forward %%%%
\subsection{Derivation of the forward pass}
\label{ssec:forward}

The quantity $\boldsymbol \alpha_t$ is defined as $\boldsymbol \alpha_t:=P(\boldsymbol \phi_t|\boldsymbol X_t)\in[0,1]^H$. Using Bayes's formula, we can write it as,
\begin{equation}
    \boldsymbol \alpha_t=\frac{p(\boldsymbol x_t|\boldsymbol \phi_t)\,P(\boldsymbol \phi_t|\boldsymbol X_{t-1})}{\sum_{\boldsymbol\phi_t^{'}}p(\boldsymbol x_t|\boldsymbol \phi_t^{'})\,P(\boldsymbol \phi_t^{'}|\boldsymbol X_{t-1})} \, .
\end{equation}
Dividing both numerator and denominator by $p(\boldsymbol x_t|\boldsymbol \phi_t)$ gives
\begin{equation}
\label{eq_at}
    \boldsymbol \alpha_t=\frac{\boldsymbol p_t}{\boldsymbol p_t+\boldsymbol r_t\,(1-\boldsymbol p_t)} \, ,
\end{equation}
where $\boldsymbol r_t$, $\boldsymbol p_t$ and $\boldsymbol \alpha_t$ correspond to the ratio of likelihood, prior and posterior probabilities of the Bayesian update respectively. One can also reformulate Equation \eqref{eq_at} by dividing the numerator and denominator by the prior. This gives rise to the well known sigmoid activation function $\sigma(x)=1/(1+e^{-x})$,
\begin{equation}
    \boldsymbol\alpha_t=\sigma\Big[\boldsymbol W^T \boldsymbol x_t + \boldsymbol b + \text{logit}(\boldsymbol p_t) \Big] \, ,
\end{equation}
where the logit function, $\text{logit}(x)=\log\big[x/(1-x)\big]$, is the inverse of the sigmoid. The prior $\boldsymbol p_t:=P(\boldsymbol \phi_t|\boldsymbol X_{t-1})$ represents the probability of having the features present at time $t$ before seeing the current observation $\boldsymbol x_t$. For a time independent prior $\boldsymbol p_t=\text{const.}$, the quantity $\text{logit}(\boldsymbol p_t)$ is also constant and can be integrated into the trainable bias $\boldsymbol b$, so that the forward pass corresponds to a hidden layer of a standard feed-forward neural network. With this probabilistic interpretation, it is therefore the time dependence of the Bayesian prior that leads to recurrence in neural networks. By assuming independent hidden features, the prior can be expanded as a function of the transition probabilities,
\begin{equation}
\label{eq_pt}
\begin{split}
    \boldsymbol p_t:&=P(\boldsymbol \phi_t|\boldsymbol X_{t-1}) \\[+3pt]
    &=P(\boldsymbol \phi_t|\boldsymbol \phi_{t-1})P(\boldsymbol \phi_{t-1}|\boldsymbol X_{t-1})\\
    &\quad\quad+P(\boldsymbol \phi_t|\neg\boldsymbol \phi_{t-1})P(\neg\boldsymbol \phi_{t-1}|\boldsymbol X_{t-1}) \\[+3pt]
    &= \boldsymbol \tau_{11}\,\boldsymbol \alpha_{t-1}+\boldsymbol \tau_{01}\,(1-\boldsymbol \alpha_{t-1}) \, .
\end{split}
\end{equation}
Using Bayes's theorem, we have thus derived a forward pass from $t=1$ to $t=T$ through the sequence, that allows the computation of $\boldsymbol \alpha_t=P(\boldsymbol \phi_t|\boldsymbol X_t)$ via a unit-wise first-order recurrence on $\boldsymbol \alpha_{t-1}$. So far, the inference on the state of the hidden features at timestep $t$ only takes the previous observations $\boldsymbol X_t=[\boldsymbol x_1,\hdots,\boldsymbol x_t]$ into account. In order to include the future observations $\boldsymbol X_{>t}=[\boldsymbol x_{t+1},\hdots,\boldsymbol x_T]$, we define a backward recursion that will smooth out the probabilities.

%%%% HMM Backward %%%%
\subsection{Derivation of HMM backward recursion}
\label{ssec:hmm_backward}

Using the relationship between joint and conditional probabilities as well as the independence of observations, we can express the desired quantity $\boldsymbol \gamma_t$ as,
\begin{equation}
\begin{split}
    P(\boldsymbol \phi_t|\boldsymbol X_T)&=\frac{P(\boldsymbol \phi_t,\boldsymbol X_T)}{P(\boldsymbol X_T)} =\frac{P(\boldsymbol X_t,\boldsymbol \phi_t)\,P(\boldsymbol X_{> t}|\boldsymbol X_t,\boldsymbol \phi_t)}{P(\boldsymbol X_{> t}|\boldsymbol X_t)\,P(\boldsymbol X_t)} \\
    &= P(\boldsymbol \phi_t|\boldsymbol X_t)\, \frac{P(\boldsymbol X_{> t}|\boldsymbol \phi_t)}{P(\boldsymbol X_{> t}|\boldsymbol X_t)} =\boldsymbol \alpha_t\,\,\boldsymbol \beta_t \, ,
\end{split}
\end{equation}
where we use the notation $\boldsymbol X_{> t}:=\boldsymbol x_{t+1},\hdots,\boldsymbol x_T$ and define $\boldsymbol \beta_t$ and its counterpart $\overline{\boldsymbol \beta_t}$ as,
\begin{equation}
\label{eq_beta}
    \boldsymbol \beta_t:=\frac{P(\boldsymbol X_{> t}|\boldsymbol \phi_t)}{P(\boldsymbol X_{> t}|\boldsymbol X_t)} \quad\text{and}\quad \overline{\boldsymbol \beta_t}:=\frac{P(\boldsymbol X_{> t}|\neg\boldsymbol \phi_t)}{P(\boldsymbol X_{> t}|\boldsymbol X_t)} \, .
\end{equation}
Let us start by expanding the numerator of $\boldsymbol \beta_t$ and use the independence of observations,
\begin{equation}
\label{eq_num1}
\begin{split}
    &P(\boldsymbol X_{>t}|\boldsymbol \phi_{t+1})\,P(\boldsymbol \phi_{t+1}|\boldsymbol \phi_t)\,+ \\
    &\,\,\,\,\,P(\boldsymbol X_{>t}|\neg\boldsymbol \phi_{t+1})\,P(\neg\boldsymbol \phi_{t+1}|\boldsymbol \phi_t) = \\[+5pt]
    &P(\boldsymbol x_{t+1}|\boldsymbol \phi_{t+1})\,P(\boldsymbol X_{> t+1}|\boldsymbol \phi_{t+1})\,P(\boldsymbol \phi_{t+1}|\boldsymbol \phi_t)\,+ \\
    &\,\,\,\,\,P(\boldsymbol x_{t+1}|\neg\boldsymbol \phi_{t+1})\,P(\boldsymbol X_{> t+1}|\neg\boldsymbol \phi_{t+1})\,P(\neg\boldsymbol \phi_{t+1}|\boldsymbol \phi_t) \, .
\end{split}
\end{equation}
The denominator of equation \ref{eq_beta} can similarly be decomposed as,
\begin{equation}
\label{eq_den1}
    P(\boldsymbol X_{>t}|\boldsymbol X_t)=P(\boldsymbol x_{t+1}|\boldsymbol X_t)\,P(\boldsymbol X_{>t+1}|\boldsymbol X_{t+1}) \, ,
\end{equation}
so that combining equations \ref{eq_num1} and \ref{eq_den1} gives,
\begin{equation}
\label{eq_btdev}
    \boldsymbol\beta_t=\frac{\boldsymbol b_1(\boldsymbol x_{t+1})\,\boldsymbol \beta_{t+1}\,\boldsymbol\tau_{11}+\boldsymbol b_2(\boldsymbol x_{t+1})\,\overline{\boldsymbol \beta_{t+1}}\,(1-\boldsymbol \tau_{11})}{P(\boldsymbol x_{t+1}|\boldsymbol X_t)} \, .
\end{equation}
We finally need to deal with the remaining denominator of equation \eqref{eq_btdev},
\begin{equation}
\begin{split}
    P(\boldsymbol x_{t+1}|\boldsymbol X_t)&=P(\boldsymbol x_{t+1}|\boldsymbol \phi_{t+1})\,P(\boldsymbol \phi_{t+1}|\boldsymbol X_t)\\
    &\quad\quad+P(\boldsymbol x_{t+1}|\neg\boldsymbol \phi_{t+1})\,P(\neg\boldsymbol \phi_{t+1}|\boldsymbol X_t) \\[+3pt]
    &=\boldsymbol b_1(\boldsymbol x_{t+1})\boldsymbol p_{t+1} \\
    &\quad\quad+\boldsymbol b_2(\boldsymbol x_{t+1})(1-\boldsymbol p_{t+1}) \, .
\end{split}
\end{equation}
By dividing the numerator and denominator by $\boldsymbol b_1(\boldsymbol x_{t+1})$, we then get the following final expression for $\boldsymbol \beta_t$,
\begin{equation}
\label{eq_bt}
    \boldsymbol \beta_t=\frac{\boldsymbol \tau_{11}\,\boldsymbol \beta_{t+1}+\boldsymbol r_{t+1}(1-\boldsymbol \tau_{11})\,\overline{\boldsymbol \beta_{t+1}}}{\boldsymbol p_{t+1}+\boldsymbol r_{t+1}\,(1-\boldsymbol p_{t+1})} \, .
\end{equation}
Similarly, one can derive that,
\begin{equation}
\label{eq_bt_}
    \overline{\boldsymbol \beta_t}=\frac{\boldsymbol \tau_{01}\,\boldsymbol \beta_{t+1}+\boldsymbol r_{t+1}\,(1-\boldsymbol \tau_{01})\,\overline{\boldsymbol \beta_{t+1}}}{\boldsymbol p_{t+1}+\boldsymbol r_{t+1}(1-\boldsymbol p_{t+1})} \, .
\end{equation}

%%%% Kalman backward %%%%
\subsection{Derivation of Kalman backward recursion}
\label{ssec:kalman_backward}

Following the approach of a Kalman smoother, a simpler backward pass can be derived by expanding $\boldsymbol \gamma_t$ on possible future states,
\begin{equation}
\begin{split}
    P(\boldsymbol\phi_{t}|\boldsymbol X_T)&=P(\boldsymbol\phi_{t}|\boldsymbol\phi_{t+1}) P(\boldsymbol\phi_{t+1}|\boldsymbol X_T) \\
    &\quad\quad+ P(\boldsymbol\phi_{t}|\neg\boldsymbol\phi_{t+1}) P(\neg\boldsymbol\phi_{t+1}|\boldsymbol X_T) \, .
\end{split}
\end{equation}
The transition probabilities need to be flipped using Bayes theorem, which gives
\begin{equation}
\begin{split}
    P(\boldsymbol\phi_{t}|\boldsymbol\phi_{t+1})&=\frac{P(\boldsymbol\phi_{t+1}|\boldsymbol\phi_{t}) P(\boldsymbol\phi_{t}|\boldsymbol X_t)}{\sum_{\phi^{'}_t}P(\boldsymbol \phi_{t+1}|\boldsymbol \phi^{'}_t)P(\boldsymbol \phi^{'}_t|\boldsymbol X_t)} \\[+5pt]
    &=\frac{\boldsymbol\tau_{11}\, \boldsymbol \alpha_{t}}{\boldsymbol\tau_{11}\,\boldsymbol \alpha_{t} + \boldsymbol\tau_{01}\,(1-\boldsymbol \alpha_{t})} =\boldsymbol\tau_{11}\frac{\boldsymbol\alpha_t}{\boldsymbol p_{t+1}}\, 
\end{split}
\end{equation}
for the first one, using the definition of the prior given in Equation \eqref{eq_pt}. Applying the same treatment to the second one then gives the following backward recursion,
\begin{equation}
\label{eq_gt}
\begin{split}
    \boldsymbol \gamma_t&=\boldsymbol \alpha_t\Bigg(\boldsymbol\tau_{11}\frac{\boldsymbol\gamma_{t+1}}{\boldsymbol p_{t+1}} + (1-\boldsymbol \tau_{11})\frac{1-\boldsymbol\gamma_{t+1}}{1-\boldsymbol p_{t+1}}
    \Bigg) \, .
\end{split}
\end{equation}

%%%% Proof %%%%
\subsection{Equivalence of the two backward recursions}
\label{ssec:proof}
Let us start with the HMM formulation of $\boldsymbol\gamma_t=\boldsymbol\alpha_t\,\boldsymbol\beta_t$. We can use Equation \eqref{eq_at} of the forward pass to rewrite $\boldsymbol\beta_t$ as,
\begin{equation}
\begin{split}
    \boldsymbol\beta_t&=\frac{\boldsymbol\alpha_{t+1}}{\boldsymbol p_{t+1}}\Big(\boldsymbol\tau_{11}\,\boldsymbol\beta_{t+1} + (1-\boldsymbol\tau_{11})\,\boldsymbol r_{t+1}\,\overline{\boldsymbol\beta_{t+1}}\Big) \\
    &=\boldsymbol\tau_{11}\frac{\boldsymbol\gamma_{t+1}}{\boldsymbol p_{t+1}} + (1-\boldsymbol\tau_{11})\frac{(1-\boldsymbol\alpha_{t+1})\overline{\boldsymbol\beta_{t+1}}}{1-\boldsymbol p_{t+1}} \, .
\end{split}
\end{equation}
By comparing with Equation \eqref{eq_gt}, we see that in order to prove that the HMM and Kalman recursions are equivalent, the following equality,
\begin{equation}
\label{eq_proof}
    1-\boldsymbol\gamma_{t+1}=(1-\boldsymbol\alpha_{t+1})\overline{\boldsymbol\beta_{t+1}}  \, ,
\end{equation}
must be satisfied $\forall t \in \{T-1,T-2,\hdots,0\}$. This can be demonstrated by induction as follows. 
\begin{proof}
We start by considering the base case $t=T-1$. Here $1-\boldsymbol\gamma_T=(1-\boldsymbol\alpha_T)\overline{\boldsymbol\beta_T}$ follows trivially from $\overline{\boldsymbol\beta_T}=1$ and $\boldsymbol\gamma_T=\boldsymbol\alpha_T$. By assuming that Equation \eqref{eq_proof} is correct for the case $n=t+1$, we must now prove that it holds for the next case $n=t$. Let us start with the left hand side $1-\boldsymbol\gamma_t$ and use the assumption for $n=t+1$ to express it as,
\begin{equation}
\label{eq_lhs}
    1-\boldsymbol\gamma_t=1-\boldsymbol \alpha_t\Bigg(\boldsymbol\tau_{11}\frac{\boldsymbol\gamma_{t+1}}{\boldsymbol p_{t+1}} + (1-\boldsymbol \tau_{11})\frac{1-\boldsymbol\gamma_{t+1}}{1-\boldsymbol p_{t+1}}
    \Bigg) \, .
\end{equation}
The transition probabilities $\boldsymbol\tau_{11}$ can be expressed as a function of $\boldsymbol\tau_{01}$ using Equation \eqref{eq_pt},
\begin{equation}
\label{eq_tautau}
    \boldsymbol\tau_{11}=\frac{\boldsymbol p_{t+1}-\boldsymbol\tau_{01}(1-\boldsymbol\alpha_t)}{\boldsymbol\alpha_t} \, .
\end{equation}
By plugging Equation \eqref{eq_tautau} into \eqref{eq_lhs}, we get that
\begin{equation}
\begin{split}
    1-\boldsymbol\gamma_t&=(1-\boldsymbol\alpha_t)\Bigg(\boldsymbol\tau_{01}\frac{\boldsymbol\gamma_{t+1}}{\boldsymbol p_{t+1}} + (1-\boldsymbol\tau_{01})\frac{1-\boldsymbol\gamma_{t+1}}{1-\boldsymbol p_{t+1}}\Bigg) \\
    &= (1-\boldsymbol\alpha_t)\,\overline{\boldsymbol\beta_t} \, .
\end{split}
\end{equation}
\end{proof}

%%%% Summary %%%%
\subsection{Implementation of the method}
\label{ssec:summary}

The ratio of the distributions $\boldsymbol b_2(\boldsymbol x_t)$ and $\boldsymbol b_1(\boldsymbol x_t)$, can be computed in advance for all timesteps and hidden features using Equations \eqref{eq_rt}. At $t=0$, $\boldsymbol \alpha_0$ is initialized with the trainable unconditional prior probability $\boldsymbol \rho_0=P(\boldsymbol \phi_0)$. A forward pass from $t=1$ to $t=T$ is then performed to compute and store the Bayesian prior $\boldsymbol p_t=P(\boldsymbol \phi_t|\boldsymbol X_{t-1})$ and posterior $\boldsymbol \alpha_t=P(\boldsymbol\phi_t|\boldsymbol X_t)$ using Equation \eqref{eq_pt} and \eqref{eq_at} respectively. Since the two backward procedures are equivalent, we use the Kalman recursion, as it is computationally simpler. At $t=T$, $\boldsymbol\gamma_T$ is initialized with $\boldsymbol\alpha_T$, and from $t=T-1$ to $t=1$, $\boldsymbol\gamma_t=P(\boldsymbol\phi_t|\boldsymbol X_T)$ is computed using Equation \eqref{eq_gt}\footnote{Code at {\scriptsize\url{https://github.com/idiap/bayesian-recurrence}}}.

%%%%%%%%%% 4.) EXPERIMENTS %%%%%%%%%%
\section{Experiments}
\label{sec:exps}

%-- Setup

Speech recognition experiments are performed on the TIMIT corpus \cite{Garofolo1993}, using the speechbrain \cite{Ravanelli2021} framework. Mel filterbank features are extracted from the waveforms and fed into two convolutional layers, followed by recurrent layers of H=512 hidden units. After two additional linear layers and a final log-softmax activation, the network outputs log-probabilities of phoneme classes. The training is done using the connectionist temporal classification (CTC) loss \cite{Graves2006} and the Adadelta optimizer \cite{Zeiler2012} for 50 epochs. Batch-normalization \cite{BN} is also used on feed-forward connections, as suggested in \cite{Ravanelli2018}.

%-- Hypotheses

Speech features entering the architecture are highly correlated. This suggests that the layer-wise recurrence of standard RNNs, which assumes interdependent hidden features, is best suited for processing them. Nevertheless, once the speech information has been processed and decorrelated, the classification of phoneme or subword representations does not require to assume the same level of correlation. As one expects a phoneme to stay in a state before transitioning to the next one, HMMs have been widely employed in ASR frameworks to process this form of information. Whilst the general aim of the experiments is to implement the derived unit-wise Bayesian recurrent units (UBRUs) with the backward recursion, and demonstrate that the mathematical predictions can be reflected practically, we also make the following hypotheses, 
\begin{enumerate}
    \item As they assume a latent space of independent hidden features, UBRUs should be best placed after layers of standard gated RNNs that can first decorrelate the highly interdependent speech features.
    \item Since future observations can already be taken into account with the analytically derived backward recursion, we expect that this method can compete with the standard bidirectional approach.
\end{enumerate}

%-- Results

We start by evaluating UBRUs on their own. We consider unidirectional and bidirectional units, with or without the backward recursion, which leads to four different models. The results are presented in Table \ref{table:results_1}. As expected, the error-rates are relatively high due to the low representational capacities of the units. Nevertheless, we observe that the derived backward recursion improves the error-rate without requiring more trainable parameters, whereas making the units bidirectional does not.

% Table 1
\begin{table}[h]
\centering
\caption{PER on TIMIT with only two layers of UBRUs.}
\label{table:results_1}
\begin{tabular}{l c c} 
 \toprule
 \textbf{Model type} & \textbf{Num. param.} & \textbf{PER}\\ [0.5ex] 
 \midrule
 Unidirectional  & 3.2M & 23.62$\%$ \\ 
 Udir. + backward & 3.2M & \textbf{22.67$\%$} \\ 
 Bidirectional & 3.7M & 24.08$\%$ \\ 
 Bidir. + backward & 3.7M & 23.27$\%$ \\ 
 \bottomrule
 \end{tabular}
\end{table}

We then test UBRUs by placing them after layers of state-of-the-art bidirectional Li-GRUs. We again find that unidirectional UBRUs with the backward recursion perform the best, as shown in Table \ref{table:results_3}, which corroborates our second hypothesis and highlights the importance of our probabilistic derivation. 

% Table 2
\begin{table}[ht]
\centering
\caption{TIMIT PER with four Li-GRU and one UBRU layers.}
\label{table:results_3}
\begin{tabular}{l c c } 
 \toprule
 \textbf{Model type} & \textbf{Num. param.} & \textbf{PER} \\ [0.5ex] 
 \midrule
 Unidirectional & 10.0M & 14.36$\%$  \\
 Udir. + backward & 10.0M & \textbf{13.96$\%$}\\
 Bidirectional & 10.3M & 14.75$\%$\\
 Bidir. + backward & 10.3M & 14.19$\%$\\
 \bottomrule
 \end{tabular}
\end{table}

By comparing with the Li-GRU baseline in Table \ref{table:results_2}, we find that adding a single unidirectional UBRU layer with the backward recursion brings the same improvement as adding another Li-GRU layer, even though the latter contains seven times more trainable parameters. In contrast, placing the UBRUs before the Li-GRUs in initial ad-hoc experiments suggested that the units were not effective at the acoustic level. This adheres to our first hypothesis that if features at that level represent phonemes and not acoustics, then the HMM-like derived UBRUs are appropriate for the classification task. 

% Table 3
\begin{table}[h]
\centering
\caption{PER on TIMIT with layers of Li-GRUs.}
\label{table:results_2}
\begin{tabular}{l c c } 
 \toprule
 \textbf{Model type} & \textbf{Num. param.} & \textbf{PER} \\ [0.5ex] 
 \midrule
 Li-GRU 4x512 & 9.8M & 14.83$\%$  \\
 Li-GRU 5x512 & 11.3M & 13.99$\%$\\
 \bottomrule
 \end{tabular}
\end{table}

For reference, we made the same experiments with cross-entropy loss inside the pytorch-kaldi \cite{Ravanelli2019, Povey2011} framework. Here again, a layer of unidirectional UBRUs with backward recursion is able to compete with a fifth Li-GRU layer, both scoring an accuracy of 14.4$\%$ compared to 14.8$\%$ for four Li-GRU layers.

\begin{comment}
Here fMLLR input features are extracted via the Kaldi recipe \cite{Povey2011}, and fed into recurrent layers, followed by a final linear layer. The networks are trained for 24 epochs using the RMSprop \cite{RMSprop} optimizer.
\end{comment}

%-- Summary

In summary, due to their correspondence with HMMs, the analytically derived unidirectional unit-wise recurrent units with a backward recursion are capable of replacing considerably larger, state-of-the-art, bidirectional, layer-wise units on the phoneme end of an ASR architecture, at an extremely low cost in terms of trainable parameters.

%%%%%%%%%% 6.) CONCLUSION %%%%%%%%%%
\section{Conclusion}

Using a probabilistic formulation of neural network components, we have analytically derived a new type of recurrent unit with a unit-wise feedback and a backward recursion. The similarity with Kalman smoothers and the forward-backward algorithm of HMMs is made explicit, and the equivalence of both approaches is proven by induction. Evaluating on a standard speech recognition task shows that the derived backward recursion gives better results compared to the conventional bidirectional approach. Moreover, adding the derived unit-wise Bayesian recurrent units after layers of larger gated RNNs is capable of considerably improving upon their performance, while only relying on a limited amount of trainable parameters, showing the importance of a probabilistic derivation.

%%%%%%%%%% ACKNOWLEDGMENTS %%%%%%%%%%
\section{Acknowledgements}
\label{sec:thanks}

This project received funding under NAST: Neural Architectures for Speech Technology, Swiss National Science Foundation grant 200021\_185010.

%%%%%%%%%% BIBILIOGRAPHY %%%%%%%%%%
\newpage
\bibliographystyle{IEEEtran}
\ninept

\bibliography{main}

\end{document}